%% file: main.tex
\documentclass[lettersize,journal]{IEEEtran}
\usepackage{amsmath,amsfonts}
\usepackage{algorithmic}
\usepackage{adjustbox}
\usepackage{algorithm}
\usepackage{array}
\usepackage{subcaption}
\usepackage{textcomp}
\usepackage{stfloats}
\usepackage{url}
\usepackage{verbatim}
\usepackage{graphicx}
\usepackage{ragged2e}
\usepackage{gensymb}
\usepackage{cite}
\usepackage{capt-of}
\usepackage{tikz}
\usetikzlibrary{arrows.meta, positioning, shadows, shapes.geometric, shapes.multipart}

\tikzstyle{startstop} = [rectangle, rounded corners, minimum width=3cm, minimum height=1.5cm,text centered, draw=black, fill=red!30]

\tikzstyle{process} = [rectangle, minimum width=3.5cm, minimum height=1cm, text centered, draw=black, fill=blue!30]
\tikzstyle{arrow} = [thick,->,>=stealth]
\newcommand{\red}[1]{\textcolor{black}{#1}}

\begin{document}

\title{\red{Digital Buildings Analysis}: 3D Modeling, GIS Integration, and Visual Descriptions Using Gaussian Splatting, ChatGPT/Deepseek, and Google Maps Platform}

\author{Kyle Gao, ~\IEEEmembership{Graduate Student Member IEEE}, Dening Lu, Liangzhi Li, Nan Chen,  Hongjie He, Linlin Xu*, \IEEEmembership{Member IEEE}, Jonathan Li, \IEEEmembership{Fellow IEEE}
\thanks{Manuscript received February xx, 2025; accepted xxxx xx 2025. Date of publication xxxx xx, 2025. This work was supported in part by the NSERC discovery grant under No. RGPIN-2022-03741. (Corresponding author: J. Li, junli@uwaterloo.ca and L. Xu, lincoln.xu@ucalgary.ca) }
\thanks{Kyle Gao, Dening Lu, and Jonathan Li (cross-appointed) are with the Department of Systems Design Engineering, University of Waterloo, Canada (e-mail: {y56gao, d62lu, junli}@uwaterloo.ca).}
\thanks{Hongjie He and Jonathan Li are with the Department of Geography and Environmental Management, University of Waterloo, Canada (e-mail: {h69he@uwaterloo.ca
, junli@}@uwaterloo.ca).}
\thanks{Nan Chen is with the School of Computer Science, Xi'an Aeronautical University, China (e-mail: chcdut@126.com).}
\thanks{Liangzhi Li is with the College of Land Engineering, Chang'an University, China (e-mail: liliangzhi@chd.edu.cn).}
\thanks{Linlin Xu is with the Department of Geomatics
Engineering, University of Calgary, Canada (e-mail: lincoln.xu@ucalgary.ca).}}

\markboth{}%
{Shell \MakeLowercase{\textit{et al.}}: 3DGS Diffusion}


\maketitle

\input{0abstract}




\input{1intro.tex}
\input{3Background_Short}
\input{4Methodology.tex}
\input{5Results.tex}

\input{6Conclusion.tex}



%

\bibliographystyle{IEEEtran}
\bibliography{References}

\newpage
 




\vfill

\end{document}

%% file: 0abstract.tex
\begin{abstract}
\red{We propose a Digital Building Analysis (DBA)}, a digital system for building-scale cloud-based data integration and data analytics. By connecting to cloud mapping platforms such as Google Map Platforms APIs, by leveraging state-of-the-art multi-agent Large Language Models data analysis using ChatGPT(4o) and Deepseek-V3/R1, and by using our Gaussian Splatting-based mesh extraction pipeline, our framework can retrieve a building's 3D model, visual descriptions, and achieve cloud-based mapping integration with large language model-based data analytics using a building's address, postal code, or geographic coordinates, and be easily extended to perform data analysis on other cloud-based data streams. 
\end{abstract}

%% file: 1intro.tex
\section{Introduction}
\input{pipeline}
In this manuscript, we present Digital Building Analysis (DBA), a framework which allows for the extraction of the 3D mesh model of a building, along with Cloud Mapping Service Integration and Multi-Agent Large Language Models (LLM) for data analysis. In the scope of this paper, we use the framework to retrieve Gaussian Splatting models and 3D mesh models. We also retrieve fundamental geocoding information, mapping information and 2D images, and perform visual analysis on the 2D images using the Multi-Agent LLM module. \red{Our framework is visualized in Fig. \ref{fig:framework}}.

Depending on need, the Google Maps Platform Integration can also retrieve local elevation maps, real-time traffic data, air quality data and access other data sources and services, which can then be analyzed. 

Our contributions are as follows.
\begin{itemize}
    \item We introduce Digital Building Analysis (DBA), a framework for extracting 3D mesh models of buildings. We integrate Cloud Mapping services for retrieving geocoding, mapping information, and 2D images.
    \item We designed a Multi-Agent Large Language Models (LLM) module for data analysis.
    \item We performed extensive visual analysis experiments of multi-view/multi-scale images of the building of interest using the LLM module. We also assessed the performance of the popular ChatGPT(4o/mini) and Deepseek-V3/R1 models.
\end{itemize}


%% file: pipeline.tex
\begin{figure*}[htpb]
\vspace{20pt}
\centering
\begin{tikzpicture}[
    node distance=1.cm and 2.cm,
    every node/.style={align=center, font=\small},
    arrow/.style={-{Latex}, thick},
    process/.style={align=center, draw, font=\small, fill=blue!30},
    process2/.style={align=center, draw, font=\small, fill=blue!15},
    outprocess/.style={align=center, rounded corners, draw, font=\small, fill=red!30}
]

\node [outprocess](googleES) {Google Earth Studio};
\node[right=of googleES, process] (gptAPI) {Multi-Agent \\ LLM Module};
\node[right=of gptAPI,process] (gee) {Google Map Platform \\  Integration};
\node[above=1cm of gptAPI] (input) {One of \{\textit{address, place name, postal code,  geographic coordinates}\}};
\node[below=of googleES,process2] (gbm) {Gaussian Building Mesh (GBM)};
\node[below left=1cm and -2.0cm of gbm] (mesh) {\hspace{-5em}3D colored mesh};
\node[right=0.4cm of mesh] (2d) {Synthesized 2D image \\ from new viewpoints};
\node[below=1.5cm of gptAPI](semantic out) {Building semantic descriptions};
\node[below=1.5cm of gee] (gis out){Cloud-based building GIS\\ information + 2D maps/images};

\draw[arrow] (input) -- (googleES);
\draw[arrow] (gee) -- (gptAPI);
\draw[arrow] (googleES) -- (gbm);
\draw[arrow] (googleES) -- (gptAPI);
\draw[arrow] (input) -- (gee);
\draw[arrow] (gbm) -- (mesh);
\draw[arrow] (gbm) -- (2d);
\draw[arrow] (gptAPI) -- (semantic out);
\draw[arrow] (gee) -- (gis out);
\end{tikzpicture}
\caption{Diagram of our Digital Twin Building framework. External modules are boxed in red. Our own tools/modules are boxed in blue. The aspects specifically presented in this paper are in dark blue. Data (both inputs and outputs) are drawn in plain text.}\label{fig:framework}
\end{figure*}
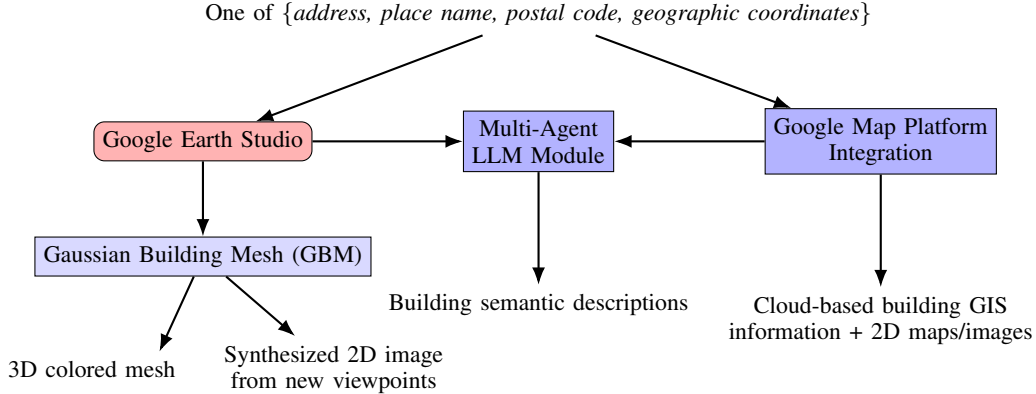

%% file: 3Background_Short.tex
\section{Background and Related Works}

\subsection{ChatGPT/Deepseek and Respective API Platforms}
Large Language Models (LLMs) are neural networks, typically Transformer-based \cite{transformer}, pre-trained on extensive, diverse text/image corpora, typically sourced from web crawls. These models, designed for Natural Language Processing (NLP), typically interpret text-based prompts and generate text-based outputs. Certain models, such as "DeepseekV3/R1" and their variants \cite{deepseekv3, deepseekr1}, support object character recognition (OCR, i.e., reading text from images). Models like "ChatGPT-4o" \cite{gpt4} and its variants additionally support full interpretation and analysis of image content.

LLMs have achieved widespread adoption since 2023. Beyond basic image and text interpretation, these models recently exhibited expert-level problem-solving in various scientific and engineering domains \cite{gpqa, math500}.

Due to their large size, LLMs often face hardware constraints for local deployment. While popular LLM providers such as OpenAI and Deepseek, provide web browser interfaces for their models, they also offer Application Programming Interfaces (APIs). These APIs enable client-side software or code to query LLMs hosted on OpenAI or Deepseek servers, facilitating large-scale data processing without requiring human-in-the-loop manipulations via browser interfaces. Unlike traditional local deep learning, which necessitates GPUs for both training and inference, API-based LLM querying requires minimal local hardware and can be deployed on devices such as mobile phones. 

\red{The platforms' Python APIs are used to initialize a client, which passes messages over the internet to the OpenAI/Deepseek servers. The client which passes messages is built with the \textit{Request} library. Messages are passed using the Hypertext Transfer Protocol (HTTP). The OpenAI/Deepseek servers then respond with the LLMs' outputs. We build our data analytics system using these APIs, allowing our system to access LLMs without hosting them on our local device.}
\input{table_models}

\subsection{Google Maps Platform}
Google Map Platform is a cloud-based mapping service and a part of Google Cloud. Its API allows the client device to connect to various cloud-based GIS, mapping, and remote sensing services hosted on the Google Cloud servers. \red{It is the Cloud Mapping Platform of choice for our research. We use the Platform's API to connect to Google Maps Platform's geocoding (location and coordinate lookup) services, as well as various data sources (2D maps, 2D orthoimages, elevation data, building polygon). The Python API client is also built using the Python \textit{Request} library with HTTP. This API client can easily be extended to include weather data, traffic data, and air quality data. However, data analysis for these modalities is outside the scope of this current research.}

Although less known in the remote sensing and GIS community than its sister application Google Earth Engine, Google Map Platform has been used in a variety of GIS research including navigation, object tracking, city modeling, image and map retrieval, geospatial data analysis for commercial and industrial applications \cite{maps1,maps2,maps3,maps4}. It is also used as part of many commercial software for cloud-based mapping integration. 


\subsection{Google Earth Studio}
Google Earth Studio \cite{google_earth_studio} is a web-based animation tool that leverages Google Earth's satellite imagery and 3D terrain data. The tool is especially useful for creating geospatial visualizations, as it is integrated with Google Earth’s geographic data. It allows for the retrieval of images from user-specified camera poses at user-specified locations. In this research, we use Google Earth Studio to retrieve 360 $\degree$ multi-view remote sensing images of a building from its address, postal code, place name, or geographic coordinates following \cite{gao_3dgs,gbm}.


%% file: table_models.tex
\begin{table*}[ht]
\centering
\captionof{table}{TABLE OF IMPORTANT DEEPSEEK AND OPENAI LLMs}
\begin{tabular}{l|l|l|l|c|c|c}
\hline
Model Name            & Model Class          & Model Type    & Image Processing & Parameters    & \begin{tabular}[c]{@{}l@{}}API call price/1M\\ Input Tokens (USD)\end{tabular} & \begin{tabular}[c]{@{}l@{}}API call price/1M\\ Output Tokens (USD)\end{tabular} \\ \hline
chatgpt4o-latest      & GPT4o                & Autoregressive & Analysis         & $\sim$ 1000+B       & 2.5                                                                & 10                                                                \\ 
gpt-4o-mini           & GPT4o Mini          & Autoregressive & Analysis         & $\sim$10's of B        & 0.15                                                               & 0.6                                                               \\ 
deepseek-chat         & Deepseek V3          & Autoregressive & OCR               & 617B          & *0.14 $\times$ 0.1*                                                    & 1.10                                                              \\ 
deepseek-reasoner     & Deepseek R1 (V3-base) & Reasoning & OCR               & 617B          & 0.14                                                               & 2.19                                                              \\ 
\textit{gpt-o1$^{1}$}                & GPT-o1 (GPT4-base) & Reasoning     & None              & $\sim$175B        & 15                                                                 & 60                                                                \\ \hline
\end{tabular}\par
\smallskip
\justifying
\noindent
Table compiled on 2025-01-31.  OpenAI models are not open-sourced, their model sizes (parameters) are estimated (B = billions, M = millions). $^1$We did not include gpt-o1 in our experiments due to cost, but we include its specifications for comparison. *The Deepseek V3 API call input token price is discounted by 90\% if input caching is used for repeated identical prompting.* \label{Tab:models}
\end{table*}

%% file: 4methodology.tex
\section{Method} \label{sec:methodology}
\subsection{Gaussian Building Mesh Extraction} \label{sec:GBM}
We use the mesh extraction procedure we introduced in December 2024 \cite{gbm}. For conciseness, the process is briefly described here, and is not benchmarked. We refer the readers to \cite{gbm} for the original implementation details and benchmark comparisons. We also refer the readers to \cite{2023gaussian_splatting} for background and theory on Gaussian Splatting.

Gaussian Building Mesh (GBM) \cite{gbm} is an end-to-end 3D building mesh extraction pipeline we recently proposed, leveraging Google Earth Studio \cite{google_earth_studio}, Segment Anything Model-2 (SAM2)\cite{SAM2} and GroundingDINO \cite{groundingdino}, and a modified \cite{2dgsp} Gaussian Splatting \cite{2023gaussian_splatting} model. The pipeline enables the extraction of a 3D building mesh from inputs such as the building's name, address, postal code, or geographical coordinates. Since the GBM uses Gaussian Splatting (GS) as its 3D representation, it also allows for the synthesis of new photorealistic 2D images of the building under different viewpoints. 

\subsection{Google Maps Platform Integration}
We use the Python client binding for Google Maps Platform Services APIs to create an integration tool to automatically retrieve the GIS and mapping information of a building. For these image analysis experiments, the data is retrieved with four API calls. The first is a Google Maps Platform Geocoding/Reverse Geocoding API call which retrieves the complete address information including geographic coordinates, entrance(s) coordinates, and building polygon mask vertex coordinates. Then, a Google Maps Platform Elevation API call is used to retrieve the ground elevation using the building's coordinates as input. Additional API calls to other Cloud Services can also be performed at this step. Finally, two API calls are made using the Google Maps Platform Static Maps API to retrieve map(s) and satellite/aerial image(s) at the desired zoom level. This process is illustrated in Figure \ref{fig:googlepipeline}. The aerial/satellite image(s) are then used as one of the inputs to our Multi-Agent LLM Module.

Our Google Map Platform Integration can easily be modified to retrieve additional data from the cloud-based mapping service by adding parallel API calls below the Geocoding/Reverse Geocoding API call. For example, if we wish to analyze real-time traffic data, we can simply perform API calls to the Traffic API. 

\red{For multi-view image analysis, from a building's address, place name, postal code, or geographic coordinates, we retrieve multi-view off-nadir images of the building of interest using Google Earth Studio or use the ones previously retrieved in the GBM module (\ref{sec:GBM}). We also retrieve top-down view aerial/satellite image(s) at different scales using the building using Google Map Platform Integration.}
\input{diagram_googlepipeline}
\subsection{Multi-Agent LLM Analysis of Multi-View/Scale Images}
The motivation of this module is to create a multi-agent LLM system to analyze the data retrieved from Google Cloud Platform Services integration. In this paper, we restrict the scope of this paper to the multi-agent content analysis of multi-view/scale images. \red{The LLM clients initialized using the API providers are stateless by default and do not retain conversation memory. Messages are passed to agents as a combination of \textit{user, assistant, or system prompts}. To create an LLM agent with conversational memory, we initialize a separate client for each agent, and we store and pass the conversation history back to the agent as \textit{assistant prompts}. We also allow each agent to construct system prompts for other agents.}

 For each retrieved image, we initiate a GPT4o/GPT4o-mini agent and prompt it to analyze the image and retrieve a set of keywords for each image. We then initiate two agents, one to aggregate the keywords from all the images of the building, and one to turn the aggregated keywords into a human-readable caption description. This process is illustrated in Fig. \ref{fig:LLMpipeline}.

\input{Diagram_LLMPipeline}

\subsection{Metrics}
Although metrics such as BLEU and CIDEr are commonly used to evaluate captioning performance, they require supervised datasets with ground truth captions. However, our images lack ground truth captions. \red{Therefore, we use the CLIP \cite{clip}, BLIP \cite{2022blip}, and PAC scores \cite{pac}, which are commonly used no-reference image captioning metrics. Readers are referred to the original papers for detailed explanations. These three scores are all based on the cosine similarity between co-embedded image and text features using a pretrained encoder.} 

\begin{equation}
\text{CLIP-BLIP-PAC Score (\%)} = 100 \frac{\mathbf{t} \cdot \mathbf{i}}{\|\mathbf{t}\| \|\mathbf{i}\|}
\end{equation}
\red{where text embedding of the caption $\mathbf{t}$, and the image embedding of the corresponding image $\mathbf{i}$ are from the respective CLIP, BLIP, PAC encoders. Additionally, the PAC score is truncated at zero and can be rescaled.}

%% file: diagram_googlepipeline.tex
\begin{figure}[htpb]
\centering
\resizebox{0.45\textwidth}{!}{
\begin{tikzpicture}[
    node distance=0.75cm and 2.cm,
    every node/.style={align=center, font=\small},
    arrow/.style={-{Latex}, thick},
    process/.style={align=center, draw, font=\small},
    process2/.style={align=center, draw, font=\small, fill=blue!30},
    outprocess/.style={align=center, rounded corners, draw, font=\small, fill=red!30}
]
\node (input) {One of \{\textit{address, proper name, postal code, geographic coordinates}\}};
\node[below=of googleES,process] (geocoding) {Geocoding/Reverse Geocoding \\API call};
\node [below= 0.5cm of geocoding](gbmout) {\textit{Address, coordinates},\\ \textit{and geometry data}};
\node[below right=0.75cm and -0.7cm of gbmout,draw, fill=white, double copy shadow] (elevation) {Elevation and \\other API call};
\node[below left=0.75cm and -0.4cm of gbmout,process] (maps) {Static Maps \\API call};
\node [below=0.4cm of elevation](elevationout) {Ground elevation data,\\ additional data of interest};
\node [below right=0.75cm and -0.75cm of maps](mapsout1) {Street map(s)};
\node [below left=0.75cm and -1.0cm of maps](mapsout2) {Satellite image(s)};

\draw[arrow] (input) -- (geocoding);
\draw[arrow] (geocoding) -- (gbmout);
\draw[arrow] (gbmout) -- (elevation);
\draw[arrow] (gbmout) -- (maps);
\draw[arrow] (maps) -- (mapsout1);
\draw[arrow] (maps) -- (mapsout2);
\draw[arrow] (elevation) -- (elevationout);

\end{tikzpicture}
}
\caption{Diagram of our Google Map Services Integration Tool.}\label{fig:googlepipeline}
\end{figure}
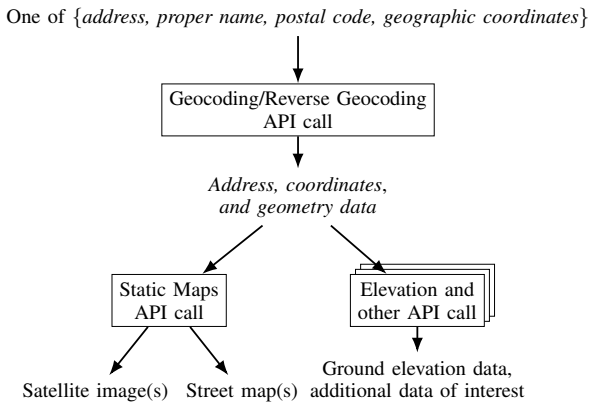

%% file: Diagram_LLMPipeline.tex
\begin{figure}[htpb]
\centering
\resizebox{0.45\textwidth}{!}{
    \begin{tikzpicture}[
node distance = 4mm and 8mm,
    I/.style = {, fill=white, minimum size=0.5cm, align=center},
     N/.style = {draw, fill=white, minimum size=1cm, align=center},
   dcs/.style = {double copy shadow, shadow xshift=2pt, shadow yshift=-2pt},
every edge/.append style = {draw, semithick, -Stealth}
                        ]
\node[I] (images)  {Multi-view\\images};
\node[N, right=of images,dcs]  (llms)       {Keyword Extraction\\LLMs};
\node[I, right= 0.5 cm of llms]  (image)       {Top-down \\
multi-scale images};
\node[draw, below=0.5 cm of llms, rectangle split, rectangle split parts=2, minimum width=3cm, minimum height=2cm] (stacked) 
    {Keyword Aggregation LLM
        \nodepart{second} Keyword-to-Caption LLM
    };
\node[I, below=of stacked] (output)  {Building caption + keywords};

\draw   (images) edge (llms)
        (image) edge (llms)
        (llms) edge (stacked)
        (stacked) edge (output);
    \end{tikzpicture}}
\caption{Diagram of Multi Agent LLM processing multi-view-multi-scale images.}\label{fig:LLMpipeline}
\end{figure}
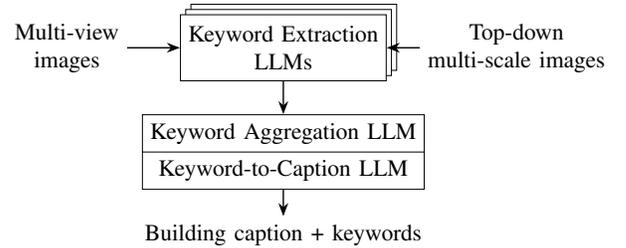

%% file: 5results.tex
\section{Experiments and Discussions}

\subsection{Experiments}
We chose seven different buildings to test our framework. These include well-known landmarks, commercial, residential, and institutional buildings. We extract 31 multi-view images in a 360$\degree$ view pose around the building of interest, which we then use in conjunction with our GBM module to create the 3D colored mesh of the building. Then we subsample six images, one every 70$\degree$, as inputs to the Multi-Agent LLM module. We also use the Google Map Platform integration to retrieve two aerial/satellite image(s), one at Google Maps zoom level 18, and one at Google Maps zoom level 19 as inputs to the Multi-Agent LLM module.


\subsubsection{End-to-end captioning}
\begin{figure}[htbp]
    \centering
    \includegraphics[width=0.35\textwidth]{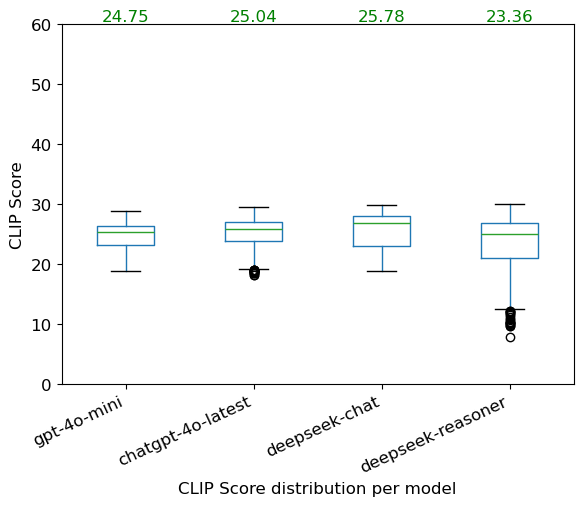}
    \caption{\red{Box plot of CLIP Scores per model (mean in green).}}
    \label{fig:clip}
\end{figure}

\begin{figure}[htbp]
    \centering
    \includegraphics[width=0.35\textwidth]{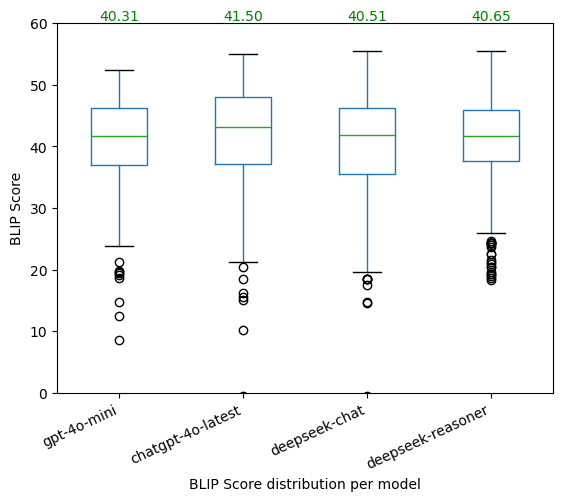}
    \caption{\red{Box plot of BLIP Scores per model (mean in green).}}
    \label{fig:blip}
\end{figure}

\begin{figure}[htbp]
    \centering
    \includegraphics[width=0.35\textwidth]{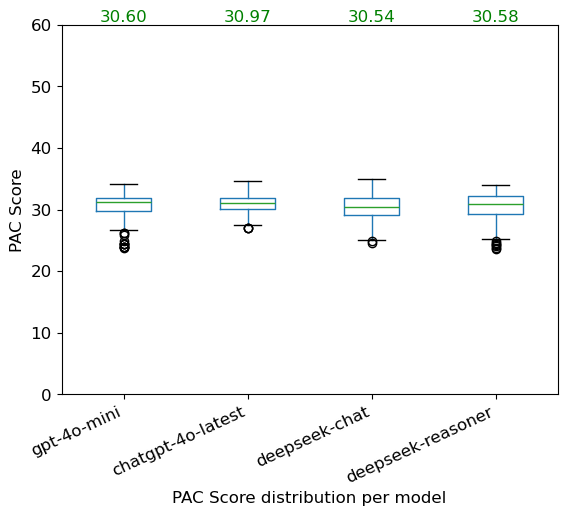}
    \caption{\red{Box plot of PAC Scores per model (mean in green).}}
    \label{fig:pac}
\end{figure}

\red{To evaluate the multi-image captioning capabilities, we run the image captioning pipeline end-to-end using \textit{gpt-4o high image resolution} API calls for keyword extraction. For each of the 7 buildings, we test 5 iterations of keyword-aggregation-caption for each of the four models: \textit{gpt-4o-mini, chatgpt-4o-latest, deepseek-chat, deepseek-reasoner}. Each test requires 10 API calls (eight calls for image keyword extraction, one for aggregation, one for caption generation), totaling 1400 API calls. We calculate CLIP, BLIP, and PAC scores for every single one of the input images. This results in $7\cdot5\cdot4\cdot8 = 1120$ triplets of scores, or 280 triplets of scores per model. The image captioning score distributions are visualized in Figs \ref{fig:clip}, \ref{fig:blip}, and \ref{fig:pac}}.

\subsubsection{Visualization}
We present a visualization of the extracted 3D model, caption, keywords, and Google Maps Platform-based information for the Perimeter Institute (PI) building scene in Fig. \ref{fig:visualizationFig}. The Perimeter Institute for Theoretical Physics is an independent research centre located at 31 Caroline St. N, Waterloo, Ontario, Canada. We show the 3D mesh and depth maps extracted from the scene, the 2D map, and the aerial image with the building's polygon at Google Maps zoom level 18, retrieved via the Google Maps Platform Static Maps API. We also plot the keywords extracted from a single view, as well as the caption generated by the Multi-Agent LLM module.
\input{Figures/visualization}

\subsection{Discussion}
\red{At each testing iteration, the generated caption differs slightly. There is some inherent variation to the captioning scores across iterations. This is why we used repeated testing with visualized as box plots to better capture this behaviour. Because we test the captions with respect to both top-down at different resolutions and angled images, there is an even larger variability in the scores than can be accounted for by differences in the generated captions. The BLIP Score and PAC Score were more sensitive to outliers. By examining these outliers, we were able to confirm that outliers occurred when combining an overly descriptive view with a viewpoint which does not contain the described image features. I.e. outliers are not model-specific as the CLIP Score in Fig \ref{fig:clip} would lead us to believe.} 

\red{The BLIP Score shows a much larger inter-quartile range than CLIP and PAC Scores, which demonstrates that it is less robust (more volatile to non-important captioning and image differences).PAC score proved to be both robust (with its small inter-quartile range) and sensitive (with its ability to detect meaningful outliers). This is consistent with the experiments \cite{pac} strongly correlating it with human assessment of captioning.}

\red{Our future research aims to leverage multi-agent LLM tools for geospatial data analysis, integrating various near real-time data sources (hourly weather, traffic, and air quality data) from Google Cloud Platform mapping services, including Google Maps Platform APIs and Google Earth Engine. In an ongoing effort, we are designing data analytics systems for these near real-time data streams to build toward digital twin systems.}

%% file: Figures/visualization.tex
\begin{figure*}[htbp]
    \centering
    \scalebox{0.80}{ 
        \begin{minipage}{\textwidth}
            \centering
            \begin{subfigure}{0.4\textwidth}
                \centering
                \includegraphics[trim=200pt 100pt 200pt 0pt, clip, width=\textwidth]{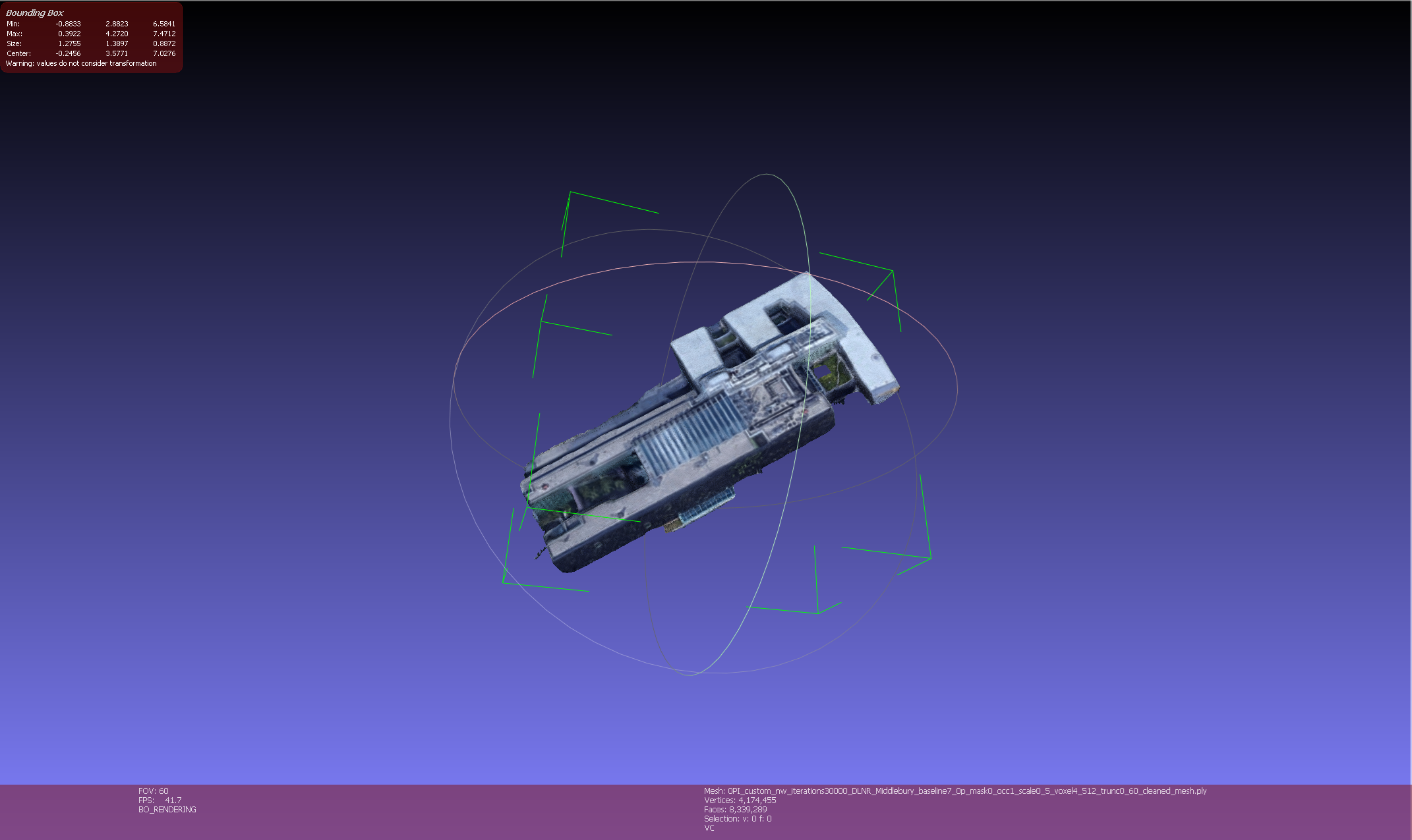}
                \label{fig:subfig1}
            \end{subfigure}
            \begin{subfigure}{0.55\textwidth}
                \centering
                \includegraphics[trim=0pt 5pt 50pt 0pt, clip, width=\textwidth]{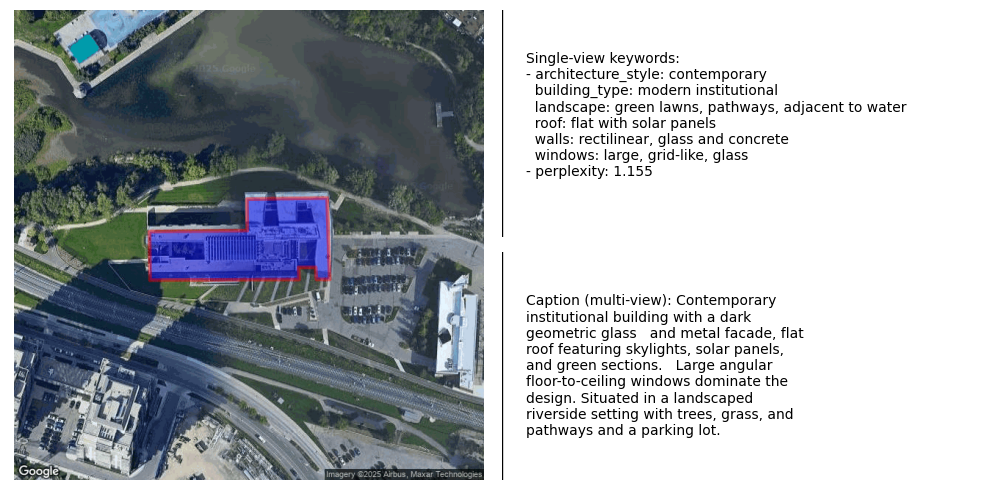}
                \label{fig:subfig2}
            \end{subfigure}
            \\
            \begin{subfigure}[b]{0.4\textwidth}
                \centering
                \includegraphics[trim=200pt 0pt 200pt 0pt, clip, width=\textwidth]{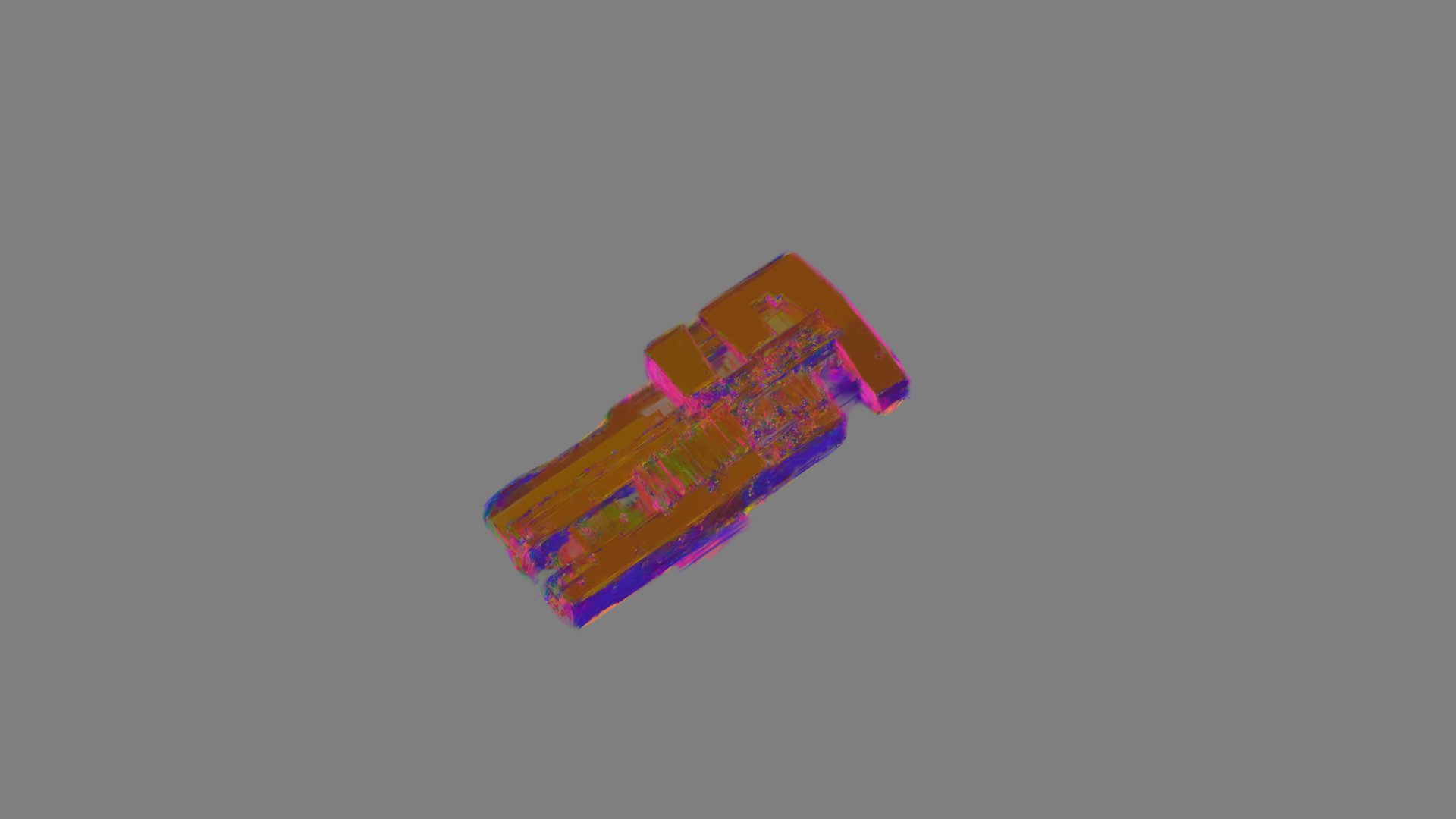}
                \label{fig:subfig3}
            \end{subfigure}
            \begin{subfigure}[b]{0.55\textwidth}
                \centering
                \includegraphics[trim=0pt 5pt 50pt 0pt, clip, width=\textwidth]{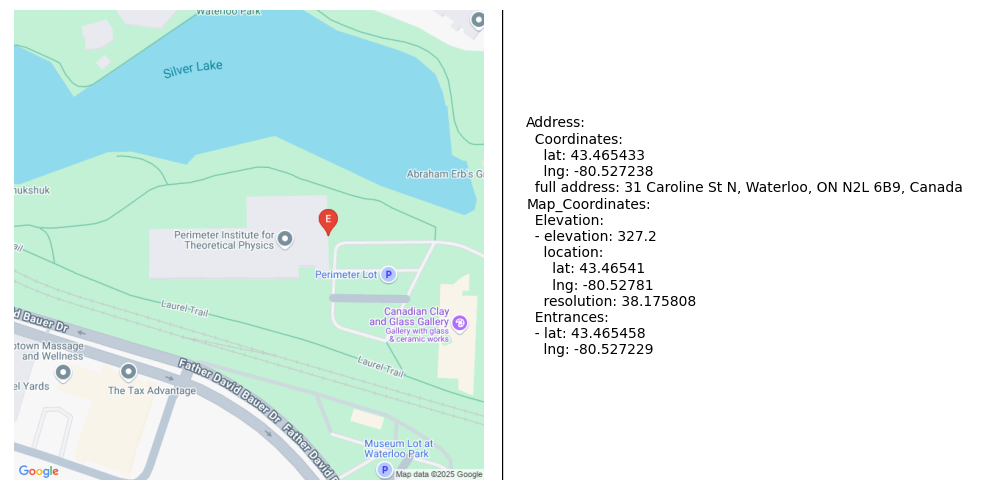}
                \label{fig:subfig4}
            \end{subfigure}
        \end{minipage}
    }
    \caption{Visualization of results. Top Left: colored 3D mesh; Bottom Left: depth map; Top Right: aerial image with keywords and captions and retrieved polygon mask. Bottom Right: retrieved map with map information. Entrance is labelled with a red place marker.}
    \label{fig:visualizationFig}
\end{figure*}

%% file: 6conclusion.tex
\section{Conclusion}
We have presented Digital Buildings Analysis, a framework for extracting the 3D mesh of a building, for connecting the building to Google Maps Platform APIs, and for Multi-Agent Large Language Models data analytics. We demonstrate this by extracting visual description keywords and captions of the building from multi-view multi-scale images of the building. The framework can also be used to process different data modalities sourced from Google Cloud Services. \red{This approach enables richer semantic understanding, seamless integration with geospatial data, and enhanced interaction with real-world structures, paving the way for advanced applications in urban analytics, navigation, and virtual environments, and can be easily extended for near real-time data streams and data analysis, building towards digital twins with real-time data analytics.}